\documentclass{article}

\usepackage[final, nonatbib]{neurips_2025}

\usepackage[utf8]{inputenc} 
\usepackage[T1]{fontenc}    
\usepackage{hyperref}       
\usepackage{url}            
\usepackage{booktabs}       
\usepackage{amsfonts}       
\usepackage{nicefrac}       
\usepackage{microtype}      
\usepackage{xcolor}         
\usepackage{amsmath}
\usepackage{amssymb}
\usepackage{mathtools}
\usepackage{amsthm}
\usepackage{bm}
\usepackage{multirow}
\usepackage{caption}
\usepackage{arydshln}  
\usepackage[capitalize,noabbrev]{cleveref}
\usepackage[table]{xcolor} 
\definecolor{lightblue}{rgb}{0.8, 0.9, 1} 
\newcommand{\gc}{\cellcolor[gray]{0.9}}
\title{Grounding Language with Vision: A Conditional Mutual Information Calibrated Decoding Strategy for Reducing Hallucinations in LVLMs}

\author{
 \textbf{Hao Fang\thanks{Equal Contribution}\ \ \textsuperscript{1}},
 \textbf{Changle Zhou$^*$\textsuperscript{1}},
 \textbf{Jiawei Kong$^*$\textsuperscript{1}},
 \textbf{Kuofeng Gao\textsuperscript{1}}, 
 \\
 \textbf{Bin Chen\textsuperscript{2}},
 \textbf{Shu-Tao Xia\thanks{Corresponding Author}\ \ \textsuperscript{1}}
\\
 \textsuperscript{1}Tsinghua Shenzhen International Graduate School, Tsinghua University,\\
 \textsuperscript{2}Harbin Institute of Technology, Shenzhen,
\\
 \small{
   \texttt{\{fangh25, kjw25, gkf21\}@mails.tsinghua.edu.cn, leliuzhe@163.com,}} \\
      \small{\texttt{chenbin2021@hit.edu.cn, xiast@sz.tsinghua.edu.cn}}
 }

\begin{document}

\maketitle

\begin{abstract}
Large Vision-Language Models (LVLMs) are susceptible to hallucinations, where generated responses seem semantically plausible yet exhibit little or no relevance to the input image. Previous studies reveal that this issue primarily stems from LVLMs' over-reliance on language priors while disregarding the visual information during decoding. To alleviate this issue, we introduce a novel Conditional Pointwise Mutual Information (C-PMI) calibrated decoding strategy, which adaptively strengthens the mutual dependency between generated texts and input images to mitigate hallucinations. Unlike existing methods solely focusing on text token sampling, we propose to jointly model the contributions of visual and textual tokens to C-PMI, formulating hallucination mitigation as a bi-level optimization problem aimed at maximizing mutual information. To solve it, we design a token purification mechanism that dynamically regulates the decoding process by sampling text tokens remaining maximally relevant to the given image, while simultaneously refining image tokens most pertinent to the generated response. Extensive experiments across various benchmarks reveal that the proposed method significantly reduces hallucinations in LVLMs while preserving decoding efficiency.
\end{abstract}
\section{Introduction}
The unprecedented breakthroughs in large vision-language models (LVLMs) \cite{zhuminigpt, liu2023visual, chen2023shikra, liu2024improved, fang2024one} have expanded their applicability across various vision-language (V+L) tasks such as autonomous driving \cite{wang2023drivemlm, cui2024survey}. Benefiting from advanced designs of model architectures and training algorithms, LVLMs trained on high-quality image-text pairs have exhibited outstanding capabilities in cross-modal alignment and complex V+L understanding.
Despite the remarkable success, the issue of hallucination continues to pose challenges to LVLMs. 
Concretely, LVLMs may generate semantically coherent yet factually incorrect contents that are entirely inconsistent with the input image \cite{chen2024halc, huo2024self, zhuang2025vasparse}. E.g., describe non-existent objects or misinterpret the attributes and relationships of visual entities within the image. 
This raises serious concerns regarding the deployment of LVLMs in real-world applications, particularly in high-risk scenarios such as medical diagnosis \cite{liu2023medical} and financial systems \cite{huang2024open}. 

To address this issue, prior work has explored several directions. One line of research
explores further fine-tuning for more fine-grained alignment \cite{liu2023mitigating, yu2024rlhf} or post-hoc analysis to correct hallucinated elements within the generated responses \cite{zhou2023analyzing, yin2024woodpecker}.
Another research stream focuses on directly modifying the token distributions in the decoding stage \cite{leng2024mitigating, wang2024mitigating, chen2024halc, huo2024self, zhuang2025vasparse, huang2024opera}. These methods employ various techniques to penalize the probabilities of hallucination-inducing tokens, thereby encouraging the generation of more faithful and reliable responses.
While these decoding-based strategies have shown practical effectiveness and efficiency
for hallucination mitigation, their designs are typically grounded in empirical findings and lack convincing theoretical foundations. Moreover, they generally fail to explicitly quantify and control the dynamic mutual relevance between the visual input and the progressively generated text, thus leading to insufficient effectiveness in certain scenarios.

\begin{figure}[t]
\centering
\includegraphics[width=0.99\linewidth]{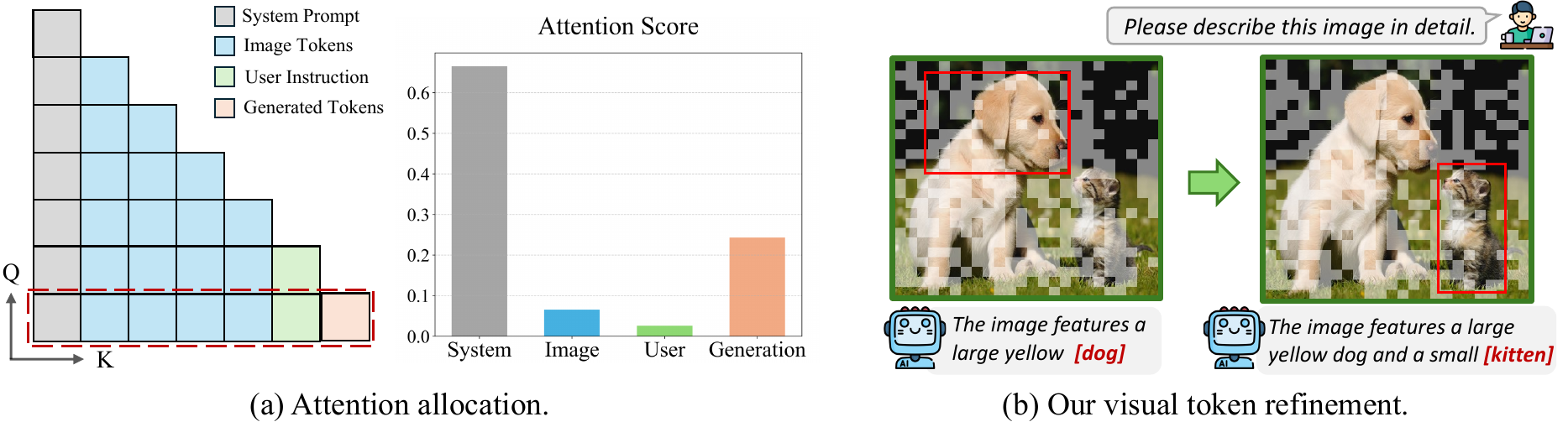}
\caption{(a) Illustration of the attention bias of LVLMs. While image tokens constitute the majority of the input tokens, they receive significantly less cumulative attention scores compared to text tokens. (b) The proposed purification mechanism when the masking ratio is 50\%. Our method promotes more reliable generation by adaptively retaining image tokens with high relevance to the ongoing response.}
\label{fig:intro}
\end{figure}

In this work, we build upon the line of decoding-based methods and investigate the issue from an information-theoretic perspective, based on which we propose a novel \textbf{C}onditional \textbf{M}utual \textbf{I}nformation-aware adaptive \textbf{V}ision-\textbf{L}anguage \textbf{D}ecoding strategy (CMI-VLD).
Specifically, previous studies \cite{yangmitigating, favero2024multi} have revealed that a key factor contributing to hallucination is LVLM's tendency to overly depend on text tokens during the autoregressive generation, with limited attention paid to the critical visual input (see Fig. \ref{fig:intro}~(a)). 
As a result, the generated text is guided more by the language priors inherent in the LLM backbone, rather than grounded in the actual visual content of the input image. This eventually leads to a low mutual dependency between the input images and the final responses, hence exacerbating the occurrence of hallucinations in LVLMs.

To mitigate this issue, we introduce conditional pointwise mutual information (C-PMI) to quantify the mutual correlation between the visual inputs and the generated texts during generation.
Correspondingly, we reformulate the hallucination mitigation objective as a vision-language mutual information maximization problem, which is further decomposed into two complementary sub-tasks that capture the respective contributions of visual and textual tokens. Based on the analysis, we derive a bi-level optimization formulation and design an effective solution that adaptively calibrates each decoding step during the generation process. 
To optimize the inner sub-problem, we calibrate the token distribution using the derived formula to prioritize tokens that exhibit strong relevance to the visual input. For the outer sub-problem, we propose an efficient visual token purifier parameterized as a learnable network, to dynamically refine image tokens that are most pertinent to the current textual context. By filtering out redundant image tokens that impair mutual information with the generated content, the proposed strategy directs the model to focus more on the key visual tokens most relevant to the ongoing response (see Fig.~\ref{fig:intro}~(b)), further enhancing the dependence of the generated text on the input image.
To summarize, our main contributions are threefold: 
 \begin{itemize}
    \item We revisit the hallucination mitigation problem in LVLMs from an information-theoretic perspective, where we reformulate it as a conditional mutual information maximization problem and introduce a novel bi-level optimization-based solution framework.  
    \item To implement this optimization, we propose an effective and efficient adaptive vision-language decoding strategy that dynamically refines the most informative visual and textual tokens to maximize the C-PMI throughout the generation process.  
    \item Extensive experimental results on multiple LVLMs such as LLaVA-1.5 across five evaluation benchmarks demonstrate the exceptional effectiveness of the proposed CMI-VLD in mitigating hallucination, significantly outperforming competitive baselines.
 \end{itemize}
 
\section{Related Work}
\textbf{Large Vision-Language Models.} Built upon advanced pre-trained LLMs \cite{touvron2023llama, touvron2023llama2, kong2025wolf, fang2025your}, LVLMs successfully bridge the gap between visual perception and linguistic reasoning \cite{ye2023mplug, li2023mimic, chen2023shikra, fang2024clip, zhang2025pre}, achieving impressive performance in generating diverse responses and tackling complex visual understanding tasks.
To incorporate visual information into the LLM backbone, LVLMs like LLaVA \cite{liu2023visual, liu2024improved} and Shikra \cite{chen2023shikra} employ linear projection layers trained by instruction fine-tuning to directly map visual features into the LLM embedding space. Meanwhile, the BLIP series \cite{li2023blip, dai2023instructblip} introduces Q-former to integrate visual tokens dynamically through gated cross-attention layers, thereby reducing redundancy in image token representations.
Benefiting from better training data, improved algorithms, and increasingly powerful LLM backbones, recent LVLMs such as LLaVA-Next \cite{li2024reqa} have demonstrated stronger multimodal understanding capabilities.
Despite the progress, LVLMs still suffer from serious hallucination problems, where the generated responses are plausible yet unfaithful or factually incorrect. Our work aims to mitigate this issue and enhance the reliability of LVLMs.

\textbf{Mitigating Hallucinations in LVLMs.} To address the critical issue, various strategies have been proposed to alleviate hallucinations from different perspectives. 
Early efforts focused on improving the multimodal alignment by training LVLMs with higher-quality data or more advanced algorithms~\cite{liu2023mitigating, yu2024rlhf, zhao2023beyond}. However, they often require additional datasets and incur substantial computational overhead, primarily due to the exhaustive instruction-tuning procedures.
In parallel, post-hoc correction methods based on auxiliary models have been explored~\cite{zhou2023analyzing, yin2024woodpecker} to filter or revise hallucinated content in the output responses. Nevertheless, these methods heavily rely on the performance of the auxiliary model and introduce extra inference overhead.

Another research line focuses on decoding-based hallucination mitigation. These methods primarily seek to construct token distributions that adaptively suppress the probabilities of hallucinated tokens~\cite{chuang2023dola, leng2024mitigating, huang2024opera, chen2024halc, wang2024mllm}. By sampling from carefully crafted distributions, these methods significantly reduce hallucinated concepts in generated responses. In addition, OPERA~\cite{huang2024open} identifies a strong correlation between hallucinations and summary tokens, and proposes to penalize the over-trust logits along with a rollback strategy. \cite{yangmitigating} conducts a modular analysis and empirically reveals that certain attention heads overly focus on textual tokens while neglecting the pivotal visual information, based on which they introduce two correction algorithms to penalize text attentions.
Among these methods, only M3ID~\cite{favero2024multi} considers theoretical aspects, yet it introduces mutual information solely to justify its vision-prompt dependency metric in contrastive decoding, without delving deeper into the key factors influencing C-PMI or exploring an effective optimization paradigm. In contrast, this paper proposes a novel multimodal adaptive decoding algorithm grounded in C-PMI, which dynamically amplifies the mutual relevance between image and text and effectively reduces hallucinations in LVLM outputs.

\section{Methodology}
This section first introduces the basic generative paradigm of LVLMs. Building on this, we propose our adaptive decoding algorithm for hallucination mitigation, \textit{i.e.}, CMI-VLD. 
Finally, we present the detailed design of a learnable predictor for visual token purification in our method.

\subsection{Preliminary}
Before delving into the proposed adaptive decoding algorithm, \textit{i.e.}, CMI-VLD, we revisit the autoregressive generation paradigm of LVLMs, which serves as the foundation for subsequent derivations. 

Given a user prompt \( x \) and an image \( v \) as input, a pre-trained LVLM \( f_{\theta}(\cdot) \) first processes the image $v$ through a vision encoder, followed by a cross-modal projection module, to generate a set of visual tokens \( v = \{v_0, v_1, \dots, v_N\} \). At decoding step $t$, the visual tokens are concatenated with the textual tokens from the instruction $x$ and the previously generated token sequence $y_{<t}$. The resulting sequence is fed into the LLM backbone of the LVLM to autoregressively predict the next token:
\begin{equation}
\begin{split}
y_t \sim {p}_{\theta}(\cdot \mid v, x, y_{<t}) = \mathrm{softmax}(f_{\theta}(\cdot \mid v, x, y_{<t})),
\end{split}
\end{equation}

where \( y_t \) is the token being sampled at current generation step \( t \). In particular, the probability of a generated sentence \( y \) of length \( l \) can be factorized as a product of conditional probabilities:
\begin{equation}
\begin{split}
{q}_{\theta}(y \mid v, x) = \prod_{t=0}^{l-1} {p}_{\theta}(y_t \mid v, x, y_{<t})=\prod_{t=0}^{l-1}\mathrm{softmax}\left(f_{\theta}\left(\cdot \mid v, x, y_{<t}\right)\right)_{y_t},
\label{equ:sentence_dis}
\end{split}
\end{equation}

where \( {q}_{\theta} \) denotes the sentence-level conditional probability distribution characterized by the LVLM $f_{\theta}(\cdot)$. This yields an appealing property for the subsequent expansion of mutual information, as the likelihood of a given text under a specific LVLM can be accurately computed by Eq.~(\ref{equ:sentence_dis}).

\subsection{The Proposed CMI-VLD}
\label{method:cmi}
To reduce hallucination-related content in the output response, we propose to strengthen the bidirectional dependency between the input image and the generated sentence by maximizing their conditional mutual information measured by the target LVLM $f_{\theta}(\cdot)$. However, standard CMI computation requires estimating the full conditional distributions of image variable \( V \) and text variable \( Y \) given the user instruction variable \( X \), which is intractable in practice due to challenges such as dimension explosion or data sparsity. 
To overcome this challenge, we adopt its pointwise formulation \cite{van2022mutual}, which balances theoretical rigor with practical feasibility, to quantify the local dependency between a specific image–text pair \( (V =v, Y =y) \), conditioned on a given instruction \( X = x \):
\begin{equation}
\begin{split}
\max_{v,y} \mathrm{C\text{-}PMI}_{\theta}(V = v, Y = y \mid X = x) = \max \left( \log \frac{p_{\theta}(v, y \mid x)}{p_{\theta}(v \mid x) \, p_{\theta}(y \mid x)} \right).
\label{equ:optim_goal}
\end{split}
\end{equation}

To achieve more effective optimization, we carefully analyze this objective from the perspectives of both visual and textual data points involved in C-PMI calculation.
Given an input image $v$, the algorithm should encourage the generation of a text $y$ that is highly aligned with the visual input $v$ to strengthen their mutual dependency.
Simultaneously, for the given text $y$, the visual input can be refined to exhibit strong relevance to $y$, hence further amplifying the mutual information between the two modalities.
As a result, the bidirectional dependency between the two variables in maximizing C-PMI naturally induces a bi-level optimization framework, which can be effectively addressed by alternately optimizing the derived inner and outer subproblems.
However, the conditional distributions in Eq. (\ref{equ:optim_goal}) can not yet be directly calculated. Based on Eq. (\ref{equ:sentence_dis}) and Bayes' Theorem, we then further expand the optimization objective as follows:
\begin{equation}
\begin{split}
\max_{v, y} \mathrm{C\text{-}PMI}_{\theta}(v, y \mid x) =\max\sum_{t=0}^{l-1} \left[ \log p_{\theta}\left(y_t \mid v, x, y_{<t}\right)- \log p_{\theta}\left(y_t \mid x, y_{<t}\right) \right].
\label{equ:optim_expansion}
\end{split}
\end{equation}

Detailed proof is in Appendix \ref{app:proof}. This formula decomposes the original objective over individual decoding steps, enabling each term to be explicitly computed using the token-level probabilities provided by the LVLM. An interesting observation is that the token distributions used in existing contrastive decoding studies~\cite{wang2024mitigating, leng2024mitigating, huo2024self, zhuang2025vasparse} can be viewed as specific variants of the optimization goal in Eq.~(\ref{equ:optim_expansion}), and thus can be naturally regarded as special cases of our framework when only the text's influence on C-PMI is considered.
Next, we concretize the solution of two interdependent subproblems from text and visual modalities to form our alternating optimization procedure:

(1) \textbf{Calibrated Distribution Sampling for Text Modality.} To optimize the text sequence $y$, Eq.~(\ref{equ:optim_expansion}) encourages us to construct an improved distribution $p_c$ to prioritize text tokens that maximize the difference between probabilities predicted with and without the visual input. However, directly applying this formulation yields unsatisfactory results since it can excessively penalize reasonable tokens in certain contexts. Inspired by \cite{leng2024mitigating, huo2024self}, we introduce a hyperparameter $\lambda$ to provide a more fine-grained control over the strength of the subtraction, which can be formally expressed as:  

\begin{figure}
\centering
\includegraphics[width=0.99\linewidth]{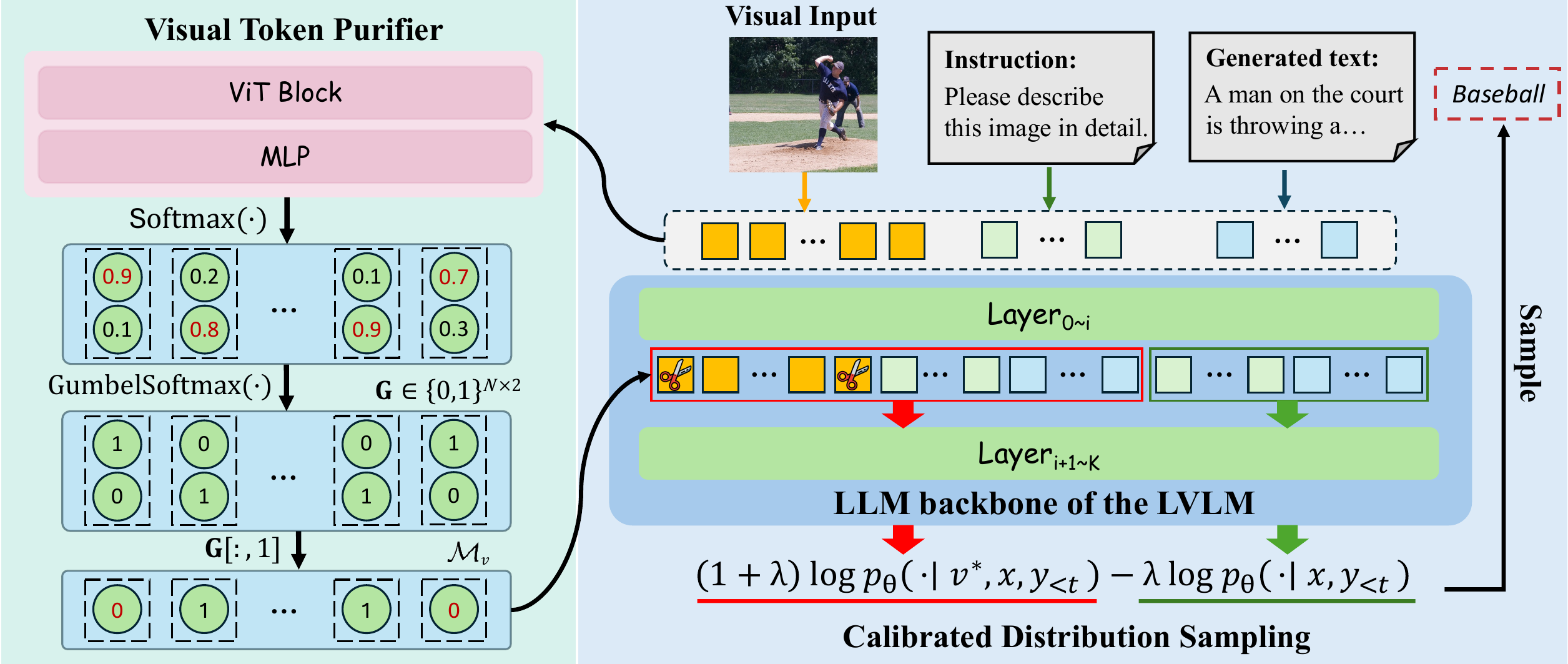}
\caption{Overview of the proposed CMI-VLD decoding. At each timestep $t$, CMI-VLD mitigates hallucination by maximizing mutual dependency between the visual input and the ongoing response through the proposed vision-language purification. Specifically, the visual token purifier first incorporates current input tokens to predict an image mask $\mathcal{M}_v$, which filters out irrelevant visual tokens to enhance C-PMI. Based on the refined visual input, a text token distribution is correspondingly constructed to penalize hallucination-related text tokens and hence guide the next-token sampling to further strengthen the dependency on the visual input.}
\label{fig:pipeline}
\end{figure}

\begin{equation}
\begin{aligned}
y_t \sim {p}_{c}(\cdot \mid v, x, y_{<t}) =  \mathrm{softmax}\Big[ (1+\lambda)f_{\theta}\left(\cdot  \mid v, x, y_{<t} \right) 
- \lambda f_{\theta}\left(\cdot \mid x, y_{<t} \right) \Big],
\label{equ:text_solution}
\end{aligned}
\end{equation}

This strategy calibrates the token distribution by urging the generation toward tokens that are more informative of the image and hence enhancing its reliance on visual content. To ensure the quality of generated sentences, we also incorporate the adaptive token truncation mechanism \cite{liu2023mitigating, leng2024mitigating} to prune the sampling space of Eq.~\ref{equ:text_solution} into a more reliable token candidate pool. 

(2) \textbf{Visual Token Refinement for Visual Modality.} Motivated by recent findings \cite{rao2021dynamicvit, chen2024image} that many image tokens in LVLMs are redundant, we propose a visual token purification mechanism that enhances C-PMI by evicting tokens considered non-informative with respect to the given text. 
In this way, the LVLM can focus more on the most critical visual tokens for improved generation. 
Moreover, we also incorporate the model’s attention scores over the visual input to identify tokens that exert a stronger influence on the LVLM’s decisions. 
Given the query vectors \( Q_i \in \mathbb{R}^{H\times n \times d_k} \) and key vectors \(K_i \in \mathbb{R}^{H\times n \times d_k} \) at the \( i \)-th LVLM layer, where \( H \) is the head number, \( n \) is the current token number, and \( d_k\) is the latent dimension, the total attention scores of an image \( v \) is calculated as:
\begin{equation}
\begin{split}
\text{Attn}_{i}(v)=\frac{1}{H}\sum_{v_j\in v}\sum_{k=0}^{H-1} A_{i}^{(k,:,:)}[-1][v_j], 
\quad \text{where}\ A_i = \mathrm{softmax}\left(\frac{Q_iK_i^\top}{\sqrt{d_k}} + \mathcal{M}_c\right),
\end{split}
\end{equation}
where \( \mathcal{M}_c \) is the causal attention mask and \( A_i \in \mathbb{R}^{H \times n \times n} \) denotes the attention matrix at the \( i \)-th layer. 
This design enables the optimizer to select visual tokens that are not only text-relevant but also highly impactful in guiding the model's predictions, boosting the effectiveness of our visual purification. Formally, the overall bi-level optimization objective can be expressed as:
\begin{equation}
\begin{aligned}
&\max_{y} \sum_{t=0}^{l-1} \Big[ (1+\lambda)\log p_{\theta}\left(y_t  \mid v^{*}, x, y_{<t} \right)
- \lambda\log p_{\theta}\left(y_t \mid x, y_{<t} \right) \Big], \\
\text{s.t.} \quad &v^{*} = \arg\max_{v} \Big[\alpha \cdot \text{Attn}_{i}(v) +  \log p_{\theta}\left(y_t \mid v, x, y_{<t} \right)
- \log p_{\theta}\left(y_t \mid x, y_{<t} \right) 
 \Big],
\label{equ:bi_level}
\end{aligned}
\end{equation}


where $\lambda$ is the aforementioned correction factor for distribution calibration and \( \alpha \) is a non-negative hyperparameter balancing the influence of C-PMI and attention scores on visual token selection. 

At decoding step $t$, we first optimize the upper subproblem by sampling $y_t$ from the distribution adjusted based on Eq.~(\ref{equ:text_solution}). 
To solve the lower subproblem, we then adaptively retain a proportion $\gamma$ of image tokens as the purified input to promote its relevance to the current textual context. Motivated by findings in \cite{chen2024image} that token sparsification at the second layer of LVLMs yields optimal performance, we utilize attention scores from this layer ($i=2$) for visual purification and start the refinement accordingly.
By alternately solving the two subproblems at each decoding step, the optimizer simultaneously samples text tokens that tightly align with current visual content and purifies informative visual tokens with strong relevance to the ongoing textual context, hence effectively amplifying C-PMI and reducing hallucination-related elements in the final response. For stable performance, we also incorporate the feature steering mechanism \cite{liu2025reducing} into our implementation.

\subsection{Visual Token Purifier for Visual Refinement}
To address the outer subproblem in Eq. (\ref{equ:bi_level}), an intuitive solution is to manually select image tokens that maximize the defined score for every decoding step. 
However, this requires repeatedly calculating token-wise scores and would incur substantial computational burdens compared to existing decoding-based approaches \cite{leng2024mitigating, wang2024mitigating}. To overcome this challenge, we propose a lightweight visual token purifier \( \mathcal{P}(\cdot) \), which consists of only a few transformer blocks and MLP layers (see Appendix \ref{sec:architecture_purifier} for details) \cite{rao2021dynamicvit}, to automatically filter visual tokens that benefit C-PMI maximization.

As illustrated in Fig.~\ref{fig:pipeline}, the purifier \( \mathcal{P}(\cdot) \) incorporates the concatenated embeddings \( \mathbf{z} = [\mathbf{z}_v, \mathbf{z}_{x}, \mathbf{z}_{y_{<t}}] \) of the image $v$ and the current text $(x, y_{<t})$ to output a probability distribution \( \bm\pi = \mathrm{softmax}\left(\mathcal{P}\left(\mathbf{z}\right)\right) \in [0, 1]^{N \times 2} \), where \( N \) is the number of visual tokens. Here, \( {\bm\pi}_{i,0} \) represents the probability of discarding the $i$-th token, and \( {\bm\pi}_{i,1} \) represents the probability of retaining it. The final visual token mask \( \mathcal{M}_v \in \{0,1\}^N \) can be then extracted via:

\begin{equation}
\begin{aligned}
\mathcal{M}_v = \left\{\operatorname*{arg\,max}_{j \in \{0, 1\}}\ \bm{\pi}_{ij}\ \middle|\ i \in \{0, 1, \dotsc, N-1\}\right\}.
\label{equ:mask_sampling}
\end{aligned}
\end{equation}

\textbf{Model Training.} The principle challenge in training the purifier lies in the non-differentiability of the \(\arg\max\) operation used for discrete token selection. To address this, we employ the Gumbel-Softmax technique with a temperature parameter \( \tau \) to enable differentiable sampling:
\label{sec:model_training}
\[
\mathbf{G} = \mathrm{Gumbel}\text{-}\mathrm{Softmax}({\bm\pi}, \tau),
\]

where the sampling output $\mathbf{G}\in\{0,1\}^{N\times2}$ containing \( N \) one-hot vectors. Since the retention probability of a visual token corresponds to the second column in $\bm\pi$, the differentiable mask \( \mathcal{M}_v \in \{0,1\}^N \) can be extracted as $\mathcal{M}_v = \mathbf{G}[:,1]$.
This approach introduces stochasticity via noise from a fixed Gumbel distribution, which enables gradients to propagate back through the probability parameters. Moreover, the temperature factor \( \tau \) helps soften the sampling distribution, thereby improving gradient stability and facilitating convergence during training.

To specify the retaining ratio $\gamma$, we introduce a Frobenius norm-based regularization term that penalizes incorrect retention rate. The overall training objective at decoding step $t$ is defined as:
\begin{equation}
\begin{aligned}
\mathcal{L}_{total} = \big(\log p_{\theta}(y_t \mid v, x, &y_{<t} ) - \log p_{\theta}(y_t \mid x, y_{<t}) \big) \\ \ + \alpha \cdot &\text{Attn}_{i}(v) + \beta\cdot\left\|\mathrm{sum}(M_v)/N - \gamma\right\|_F,
\label{equ:purifier_training}
\end{aligned}
\end{equation}
where \( \beta \) is a weight coefficient controlling the regularization strength of the reduction ratio, $\left\|\cdot\right\|_F$ denotes the F-norm of a matrix, and \(\mathrm{sum}(\cdot) \) represents the summation operation.

By iteratively updating the network using the loss function in Eq.~(\ref{equ:purifier_training}) on paired image-text data, the purifier learns to dynamically identify visual tokens that effectively contribute to mutual information maximization, which further enhances the informativeness of the visual input while discarding those redundant and distracting visual tokens. Furthermore, our method preserves the decoding efficiency despite introducing an additional network, as the purifier module is lightweight and the removal of non-essential visual tokens helps reduce the overall inference cost (see Sec .~\ref{inference_time}).

\section{Experiments}

\subsection{Experimental Setup}
\textbf{Models and Baselines.} We align with \cite{huo2024self} and choose four representative LVLMs for evaluation, including InstructBLIP \cite{dai2023instructblip}, Shikra \cite{chen2023shikra}, LLaVA-1.5 on the 7B scale \cite{liu2024improved}, and LLaVA-NeXT \cite{li2024reqa} on the 8B scale.
We conduct a comprehensive evaluation of the proposed CMI-VLD on a range of state-of-the-art (SOTA) baselines, including Sampling (Top-p=1), Greedy, VTI \cite{liu2025reducing}, VCD \cite{leng2024mitigating}, ICD \cite{wang2024mitigating}, HALC \cite{chen2024halc}, OPERA \cite{huang2024opera}, SID \cite{huo2024self}, and VASparse \cite{zhuang2025vasparse}. Following SID, we implement the proposed method under both sampling and greedy decoding settings. It is worth noting that HALC, OPERA, and VASparse adopt the more flexible and stronger beam search strategy, which may raise fairness concerns as it retains a broader set of promising candidate paths during decoding. Nevertheless, our CMI-VLD still consistently outperforms these methods.
\begin{table}[t]
  \centering
  \setlength{\tabcolsep}{8pt}
  \caption{Comparison of the proposed CMI-VLD with SOTA baselines on the CHAIR metric. We evaluate the performance on MSCOCO. The $^{\dagger}$ indicates decoding strategies based on beam search.}
    \resizebox{0.88\linewidth}{!}{\begin{tabular}{l|cc|cc|cc|cc} \toprule
    \multicolumn{1}{l|}{\multirow{2}[0]{*}{Method}} & \multicolumn{2}{c|}{\textbf{LLaVA-1.5}} & \multicolumn{2}{c|}{\textbf{InstructBLIP}} & \multicolumn{2}{c|}{\textbf{Shikra}} & \multicolumn{2}{c}{\textbf{LLaVA-Next}} \\
      & \multicolumn{1}{c}{$C_S\downarrow$} & \multicolumn{1}{c|}{$C_I\downarrow$} & \multicolumn{1}{c}{$C_S\downarrow$} & \multicolumn{1}{c|}{$C_I\downarrow$} & \multicolumn{1}{c}{$C_S\downarrow$} & \multicolumn{1}{c|}{$C_I\downarrow$} & \multicolumn{1}{c}{$C_S\downarrow$} & \multicolumn{1}{c}{$C_I\downarrow$} \\ \midrule
\textit{Sampling} & 52.2  & 15.8  & 55.0  & 25.3  & 56.2  & 15.8  & 34.8  & 9.4 \\
ICD    & 51.0  & 15.2  & 64.0  & 20.2  & 56.6  & 15.5  & 33.4  & 8.7 \\
VCD   & 50.4  & 15.6  & 57.6  & 19.2  & 56.2  & 15.5  & 36.0  & 9.3 \\
VTI   & 37.2  & 11.4  & \textbf{49.2} & 21.9  & 47.0  & 14.1  & 32.2  & 7.8 \\
SID   & 49.2  & 14.5  & 58.0  & 18.7  & 54.4  & 14.4  & 39.4  & 9.9 \\
\gc CMI-VLD  & \gc\textbf{30.2} &\gc  \textbf{9.3} & \gc  51.0  & \gc \textbf{16.1} & \gc  \textbf{38.2} & \gc  \textbf{10.1} & \gc \textbf{30.6} & \gc \textbf{7.6} \\ \midrule
\textit{Greedy} & 45.0  & 13.5  & 52.2  & 21.8  & 54.8  & 15.8  & 31.6  & 8.2 \\
ICD   & 44.8  & 12.8  & 48.8  & 14.1  & 55.0  & 14.0  & 32.8  & 9.1 \\
VCD   & 49.4  & 14.0  & 46.6  & 13.3  & 55.8  & 15.3  & 36.8  & 9.4 \\
HALC$^\dagger$   & 33.2	& 10.3 & 61.4 &	20.0 & 55.4 & 14.7  & 36.7  & 9.5 \\
OPERA$^\dagger$ & 39.4  & 10.3  & 48.2  & 13.8  & 36.8  & 11.7  & 33.6  & 8.3 \\
VTI   & 30.6  & 10.1  & 48.3  & 20.7  & 44.6  & 13.7  & 30.1  & 7.6 \\
SID   & 42.8  & 12.1  & 56.2  & 15.8  & 51.2  & 13.6  & 38.0  & 8.9 \\
VASparse$^\dagger$   & 49.6	& 14.2 & 53.6 & 14.9 & 51.6 & 14.8  & 33.6   & 9.1 \\
\gc CMI-VLD  & \gc \textbf{29.9} & \gc \textbf{8.9} & \gc \textbf{43.2} & \gc \textbf{12.9} & \gc \textbf{30.6} & \gc \textbf{8.9} & \gc \textbf{27.2} & \gc \textbf{6.8} \\ \bottomrule
    \end{tabular}}
    \vspace{-0.8em}
  \label{tab:chair}%
\end{table}%
\begin{figure}[t]
\centering
\includegraphics[width=\linewidth]{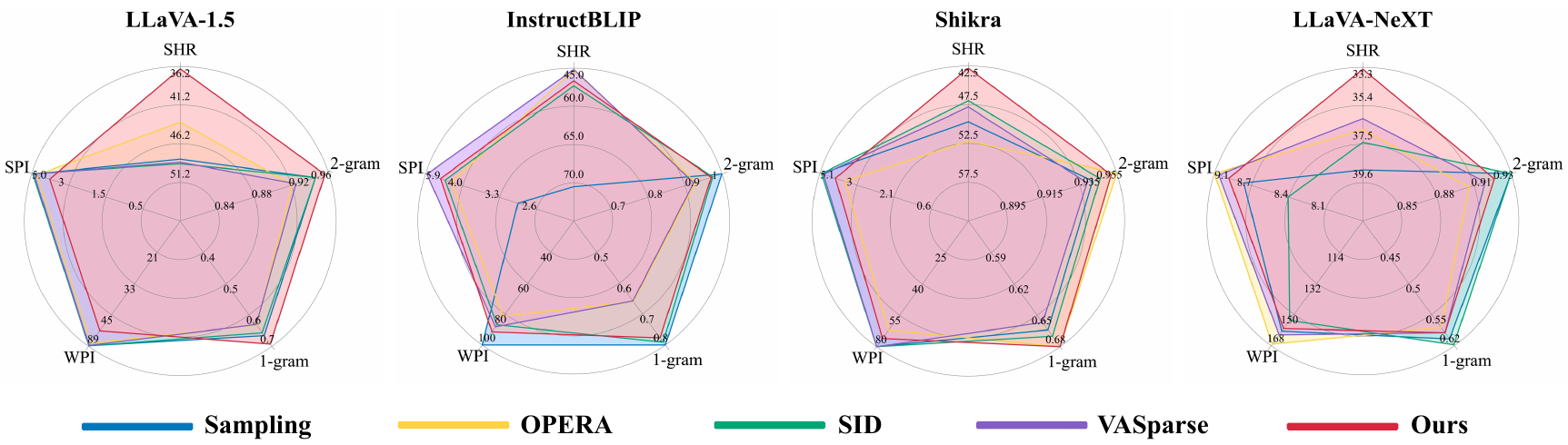}
\caption{GPT-4o assisted benchmark. We calculate the Sentence-level Hallucination Ratio (SHR) as the major metric for hallucination degree, along with 1\&2-gram, the number of sentences per image (SPI), and the number of words per image (WPI). A larger radar area indicates better performance.}
\vspace{-0.2em}
\label{fig:gpt_4o}
\end{figure}

\textbf{Evaluation Benchmarks.} Following the evaluation protocol in \cite{huo2024self, wang2024mllm}, we analyze our CMI-VLD across five widely-used benchmarks:
(1) the CHAIR metric \cite{rohrbach2018object} on the MSCOCO dataset \cite{lin2014microsoft} that measures object hallucinations; 
(2) a GPT-4 assisted evaluation \cite{zhao2023beyond}, where we adopt the advanced GPT-4o \cite{achiam2023gpt} to detect more fine-grained hallucinations and compute the Sentence-level Hallucination Ratio (SHR);
(3) Polling-based Object Probing Evaluation (POPE) \cite{li2023evaluating}, another object hallucination evaluation also conducted on MSCOCO;
(4) Multimodal Large Language Model Evaluation (MME) \cite{liang2024survey}, a general-purpose benchmark for assessing multimodal capabilities; and
(5) MMBench \cite{liu2024mmbench}, which includes multiple-choice questions designed to evaluate visual perception and reasoning.

\textbf{Implementation Details.} We set \( \alpha \) = $1\times10^2$ and \( \beta \) = $5\times10^2$ in Eq.~(\ref{equ:purifier_training}) during the training of the purifier for all LVLMs. To initiate the refinement process, we adopt Layer \( i=2 \) for LLaVA-1.5, Shikra, and LLaVA-NeXT and \( i=4 \) for InstructBLIP. 
In Sec.~\ref{exp:ablation}, we explore the contrastive strength $\lambda$ ranging from $0$ to $0.9$. 
For all experiments, we set \textit{max new tokens} as 512 for evaluation. Note that CMI-VLD is compatible with the feature steering mechanism in VTI \cite{liu2025reducing}, which is then incorporated in our implementation for stable and enhanced performance. 
More details are in Appendix \ref{sec:exp_details}.

\subsection{Performance Evaluation}
\textbf{CHAIR Evaluation.} Following previous studies \cite{huang2024opera, huo2024self, chen2024halc}, we query LVLMs with the input prompt "\texttt{Please describe this image in detail.}" using 500 images randomly sampled from the validation set of MSCOCO. By dynamically amplifying the mutual relevance between visual inputs and generated texts, the proposed method achieves remarkable improvements over SOTA baselines on different LVLMs, as observed in Table \ref{tab:chair}. \textit{E.g.}, a notable improvement of 7\% and 2.1\% in $C_S$ and $C_I$ for \textit{Sampling} on the LLaVA-1.5 model. Notably, some baselines even exacerbate hallucination content compared to standard decoding strategies in some cases. In contrast, the proposed CMI-VLD consistently reduces both sentence-level and instance-level object hallucinations in the final responses.

\begin{table}[t]
  \centering
  \caption{Comparison of the proposed CMI-VLD with SOTA baselines on the POPE metric. The $^{\dagger}$ indicates decoding strategies based on beam search.}
    \resizebox{\linewidth}{!}{\begin{tabular}{llcccccc} \toprule
    \multirow{2}[0]{*}{Model} &   \multirow{2}[0]{*}{Method}    & \multicolumn{2}{c}{Random} & \multicolumn{2}{c}{Popular} & \multicolumn{2}{c}{Adversarial} \\ \cmidrule(lr){3-4} \cmidrule(lr){5-6} \cmidrule(lr){7-8}
          &  & \multicolumn{1}{l}{Accuracy} & \multicolumn{1}{l}{F1 score} & \multicolumn{1}{l}{Accuracy} & \multicolumn{1}{l}{F1 score} & \multicolumn{1}{l}{Accuracy} & \multicolumn{1}{l}{F1 score} \\ \midrule
    \multirow{15}[0]{*}{LLaVA-Next} & \textit{Sampling} & 82.53\% & 79.19\% & 81.57\% & 78.31\% & 80.30\% & 77.16\% \\
          & ICD   & 82.77\% & 79.57\% & 81.97\% & 78.81\% & 81.03\% & 77.95\% \\
          & VCD   & 83.67\% & 80.80\% & 82.17\% & 79.40\% & 80.90\% & 78.25\% \\
          & VTI   & 82.70\% & 79.45\% & 81.43\% & 78.25\% & 80.10\% & 77.05\% \\
          & SID   & 84.67\% & 82.20\% & 83.57\% & 81.16\% & 81.60\% & 80.27\% \\ \
          & \gc CMI-VLD &\gc \textbf{85.17\%} &\gc \textbf{83.02\%} & \gc\textbf{84.10\%} &\gc \textbf{82.03\%} &\gc \textbf{82.30}\% &\gc \textbf{80.40\%} \\ \cmidrule{2-8}
          & \textit{Greedy} & 83.40\% & 80.32\% & 82.60\% & 79.55\% & 81.77\% & 78.77\% \\
          & ICD   & 83.47\% & 80.41\% & 82.60\% & 79.56\% & 81.90\% & 78.91\% \\
          & VCD   & 84.43\% & 81.85\% & 83.30\% & 80.77\% & 82.33\% & 79.88\% \\
          & HALC$^\dagger$  &   83.34\% & 80.36\% &	82.33\% &	79.48\% &	81.40\% &	78.92 \%  \\
          & OPERA$^\dagger$ & 83.50\% & 80.46\% & 82.70\% & 79.69\% & 81.87\% & 78.91\% \\
          & VTI   & 84.70\% & 82.09\% & 83.67\% & 81.11\% & 82.90\% & 80.40\% \\
          & SID   & 84.97\% & 82.53\% & 83.93\% & 81.56\% & 82.97\% & 80.67\% \\
          & VASparse$^\dagger$ &  83.47\% & 80.52 \% & 82.24 \% &	79.69\% &	81.33\% &	78.88 \%       \\
          & \gc CMI-VLD &\gc \textbf{86.43\%} &\gc \textbf{84.52\%} &\gc \textbf{85.07\%} &\gc \textbf{83.22\%} &\gc \textbf{83.90\%} & \gc\textbf{82.14\%} \\ \midrule
    \multirow{15}[0]{*}{InstructBLIP} & \textit{Sampling} & 82.03\% & 81.30\% & 78.77\% & 78.66\% & 76.37\% & 76.81\% \\
          & VTI   & 83.50\% & 82.01\% & 80.83\% & 79.70\% & 79.13\% & 78.29\% \\
          & ICD   & 83.20\% & 82.29\% & 79.87\% & 79.51\% & 77.63\% & 77.74\% \\
          & VCD   & 83.43\% & 82.49\% & 79.70\% & 79.36\% & 77.53\% & 77.65\% \\
          & SID   & 85.43\% & 84.81\% & 82.43\% & 82.24\% & 79.47\% & 79.84\% \\
          & \gc CMI-VLD &\gc \textbf{86.33\%} &\gc \textbf{85.41\%} &\gc \textbf{84.60\%} &\gc \textbf{83.87\%} &\gc \textbf{81.57\%} &\gc \textbf{81.29\%} \\ \cmidrule{2-8}
          & \textit{Greedy} & 87.27\% & 85.91\% & 84.87\% & 83.72\% & 82.97\% & 82.04\% \\
          & ICD   & 87.23\% & 85.82\% & 84.90\% & 83.68\% & \textbf{83.13}\% & 82.11\% \\
          & VCD   & 86.73\% & 85.30\% & 84.37\% & 83.16\% & 82.47\% & 81.49\% \\
          & HALC$^\dagger$  & 87.30\% & 85.96\% & 84.83\% & 83.70\% & 83.00\% & 82.08\%  \\
          & OPERA$^\dagger$ & 87.53\% & 86.26\% & 85.07\% & 84.00\% & 83.07\% & 82.24\% \\
          & VTI   & 85.73\% & 83.86\% & 84.13\% & 82.36\% & 82.50\% & 80.89\% \\
          & SID   & 88.10\% & 87.15\% & 85.87\% & 85.10\% & 82.90\% & 82.52\% \\
          & VASparse$^\dagger$ & 87.33\% & 86.00\% & 84.87\% & 83.74\% & 83.00\% & 82.09\% \\
          &\gc CMI-VLD & \gc\textbf{88.37\%} &\gc \textbf{87.50\%} & \gc\textbf{86.10\%} &\gc \textbf{85.40\%} &\gc 82.87\% &\gc \textbf{82.64\%} \\ \bottomrule
    \end{tabular}}
  \label{tab:POPE}%
  \vspace{-1em}
\end{table}%

\textbf{GPT-4o Assisted Evaluation.}
While CHAIR is a reliable evaluation metric widely adopted in previous studies, it is limited within the scope of object hallucinations and fails to identify other types, such as attribute, relational, and positional hallucinations. To more comprehensively evaluate the effectiveness of our method, we introduce the GPT-assisted benchmark \cite{zhao2023beyond}, which uses the object-level descriptions in the Visual Genome dataset \cite{krishna2017visual} as ground-truth, to judge more fine-grained hallucinations assisted by the advanced GPT-4o. Figure \ref{fig:gpt_4o} demonstrates that the proposed CMI-VLD significantly outperforms SOTA baselines across four LVLMs. Compared with competitive baselines, we achieve a relative improvement of 15.89\% for LLaVA-1.5 in the hallucination metric SHR while maintaining text fluency. We also note that our method reduces the length of generated texts to some extent, which can be caused by the removal of hallucinated sentences \cite{huang2024opera}.

\textbf{POPE Evaluation.}
The POPE metric also focuses on object hallucinations by using a prompt "\texttt{Is there a <object> in the image?}" to query LVLMs for answering a yes/no question. We report the results of the accuracy and F1 score in Table \ref{tab:POPE}. The quantitative results reveal that our method generally performs best across the three split datasets. 
Notably, in the POPE evaluation, where responses are typically short and follow fixed patterns such as "\texttt{Yes, there is a <object> in the image.}", the evaluation primarily hinges on the first one or few tokens (\textit{i.e.}, Yes or No) ~\cite{huang2024opera}. Consequently, our CMI-VLD may not fully exhibit its effectiveness in this constrained setup, as it is designed to dynamically adjust decoding over the entire generation rather than concentrating solely on the initial tokens. Nevertheless, our method still achieves notable improvements over competitive baselines.
Due to the page limit, results on more LVLMs are provided in Appendix \ref{sec:more_exp}.

\textbf{MME and MMBench Evaluations.} Apart from the above benchmarks tailored for hallucination evaluation, we additionally test on two popular LVLM benchmarks, \textit{i.e.}, MME \cite{cui2024survey} and MMBench \cite{liu2024mmbench}, to systematically analyze their various capability dimensions. MME provides a suite of fine-grained, image-grounded multiple-choice questions across various categories. We follow SID and report the overall perception score covering 10 sub-tasks, such as object existence and fine-grained recognition. MMBench is another large-scale bilingual benchmark consisting of over 3,000 curated multiple-choice questions. We compute the LVLM’s average score across 20 multimodal tasks, such as attributes, logical reasoning, and coarse/fine-grained perception, to comprehensively evaluate its capabilities. 
As observed, CMI-VLD not only reduces the hallucinated contents but also enhances diverse capabilities of MLLMs, bringing remarkable improvements over the default decoding methods. These results underscore CMI-VLD as a reliable and practical strategy for hallucination mitigation.


\begin{table}[t]
  \centering
  \caption{Comparison of the proposed CMI-VLD with SOTA baselines on LVLM benchmarks. The $^{\dagger}$ indicates beam search-based methods, while other methods adopt the same \textit{Greedy} decoding.}
    \resizebox{\linewidth}{!}{\begin{tabular}{lccccccccc} \toprule
    Benchmarks & \multicolumn{1}{c}{\textit{Greedy}} & \multicolumn{1}{c}{ICD} & \multicolumn{1}{c}{VCD} & \multicolumn{1}{c}{HALC$^\dagger$} & \multicolumn{1}{c}{OPERA$^\dagger$} & \multicolumn{1}{c}{VTI} & \multicolumn{1}{c}{SID} & \multicolumn{1}{c}{VASparse$^\dagger$} & \multicolumn{1}{c}{CMI-VLD} \\ \midrule
    MME   & 1465.11 & 1432.43 & 1472.57 &  1473.43 & 1471.37 & 1435.35 & 1467.05 &  1466.61     & \textbf{1481.17} \\
    MMBench & 64.86 & 64.26 & 64.26 &  57.90     & 64.78 & 64.43 & 64.26 &  64.78    & \textbf{65.12} \\ \bottomrule
    \end{tabular}}
  \label{tab:mme_mmbench}%
\end{table}%
\begin{figure*}[t]
\begin{minipage}[t]{0.34\linewidth}
  \centering
  \includegraphics[width=\textwidth]{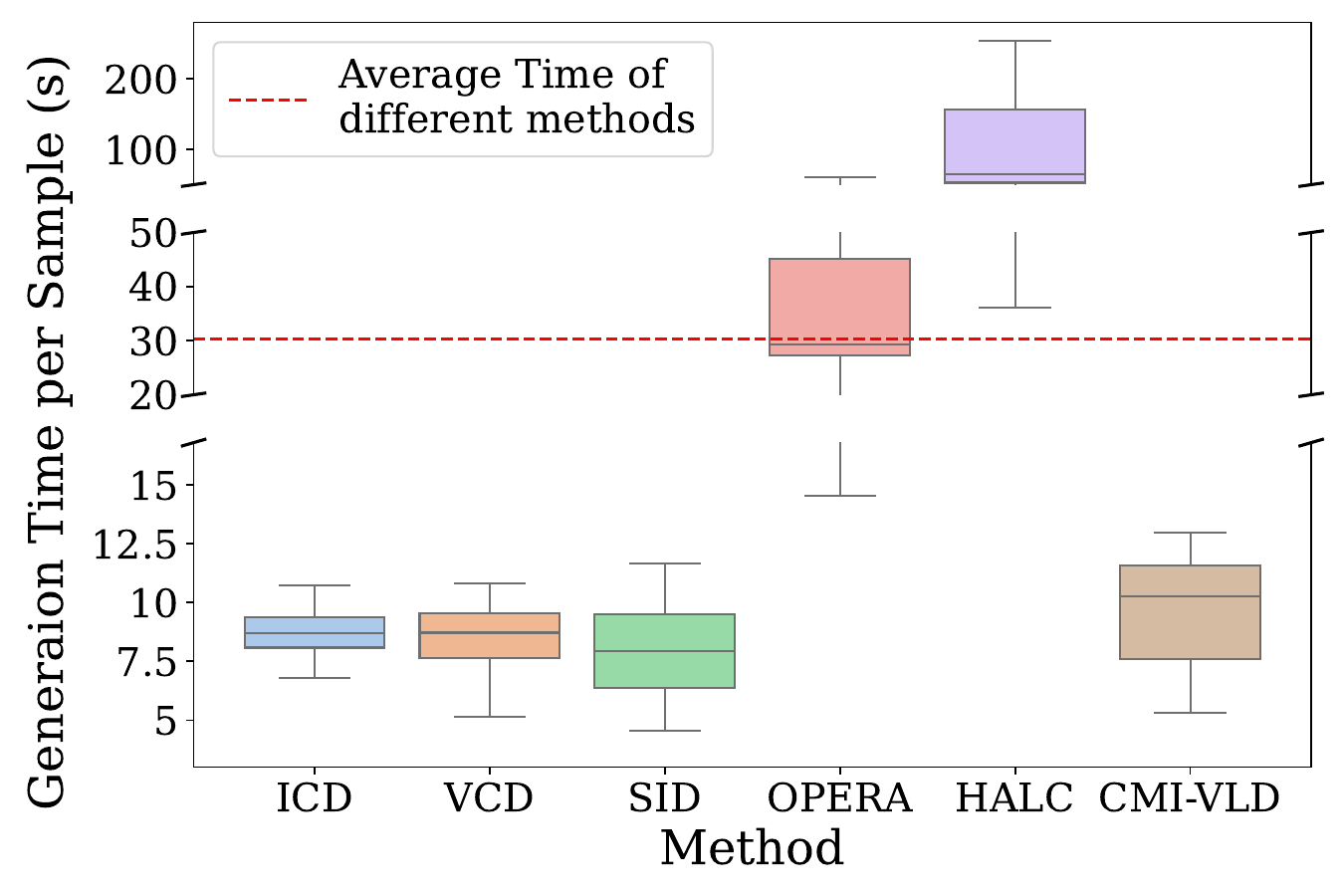}
  \centering
  \captionof{figure}{Generation time per sample of different methods.}
  \label{fig:generation_time}
\end{minipage}
\begin{minipage}[t]{.01\linewidth}
\quad
\end{minipage}
\begin{minipage}[t]{.65\linewidth}
  \centering
  \includegraphics[width=\textwidth]{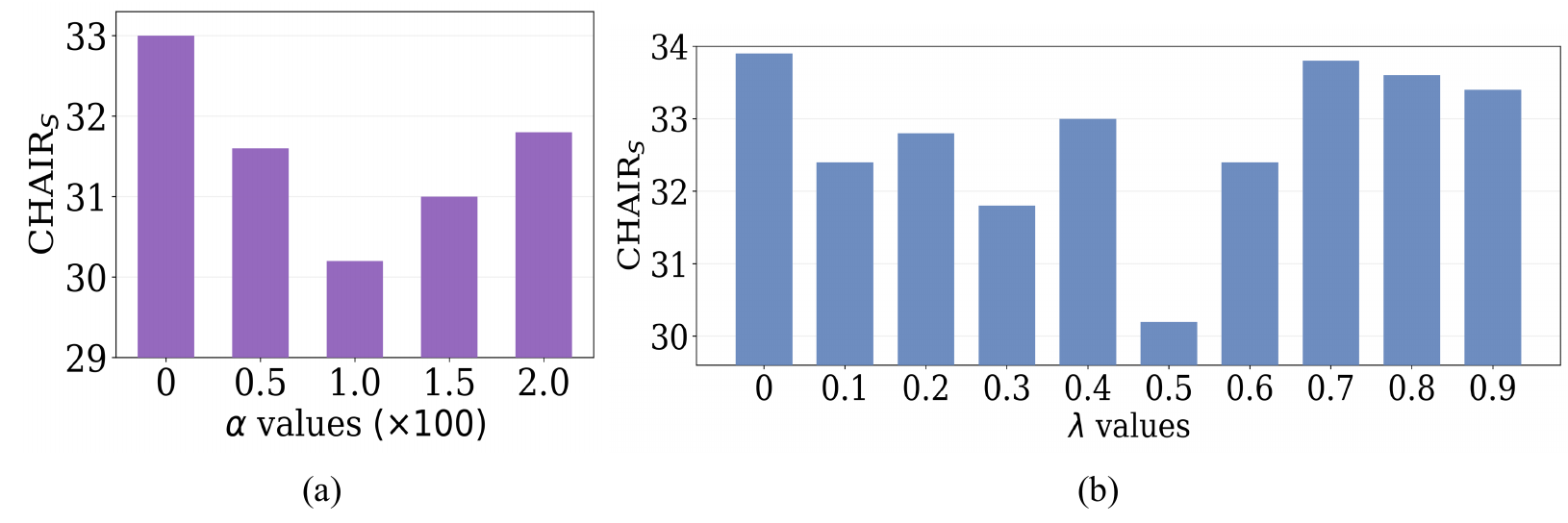}
  \captionof{figure}{CHAIR$_S$ results of the proposed CMI-VLD under varying values of hyperparameters $\alpha$ and $\lambda$.}
  \label{fig:ablation_study}
\end{minipage}

 \end{figure*}

\textbf{Inference Time Analysis.} Since our method introduces an additional visual token purifier for effective visual refinement, it is crucial to assess its influence on the overall computational efficiency. Following \cite{yangmitigating}, we calculate the generation time per response based on LLaVA-1.5 to assess computing burdens. Figure \ref{fig:generation_time} reveals that the proposed CMI-VLD achieves satisfactory decoding efficiency, introducing negligible computational overhead. This is primarily attributed to the lightweight architecture of the visual purifier and the removal of redundant visual tokens that would incur significant computational overhead, demonstrating that CMI-VLD effectively balances performance and efficiency.
\label{inference_time}

\subsection{Ablation Study}
\label{exp:ablation}
Next, we provide ablation analysis regarding several critical hyperparameters. More ablation analysis about the retaining ratio and the effectiveness of the proposed techniques is presented in Appendix \ref{sec:more_exp}.

\textbf{The effect of varying loss parameter $\alpha$.} The value of $\alpha$ is a critical factor as it adjusts the contribution of the attention scores during purifier training. We then evaluate the performance under various values of $\alpha$ in Figure \ref{fig:ablation_study} (a). The performance gains observed when comparing to $\alpha=0$ suggest that incorporating Attn$_{i}(\cdot)$ enhances the effectiveness of the visual purifier. Moreover, the results indicate that $\alpha=1\times10^2$ yields the best performance, and is therefore adopted in our main experiments.

\textbf{The effect of calibration intensity $\lambda$.} During decoding, the hyperparameter $\lambda$ plays a pivotal role in regulating the strength of distribution calibration. We present the CHAIR results under varying $\lambda$ in Figure \ref{fig:ablation_study} (b). The proposed method reaches optimal performance when $\lambda=0.5$, hence we adopt it as the default setup. Moreover, we emphasize that a properly selected range of positive values of $\lambda$ yields significant improvements over the $\lambda=0$ setup, validating our distribution calibration strategy.


\section{Conclusion}
In this work, we first revisit the key reason for hallucination in LVLMs, based on which we introduce conditional mutual information as a theoretical foundation to enhance the mutual dependency between visual input and generated text. To strengthen this cross-modal association, we propose a novel CMI-aware bi-level optimization framework, which is efficiently and effectively solved via a carefully designed vision-language decoding strategy. 
Through extensive experiments across multiple LVLMs and evaluation benchmarks, we demonstrate the superiority of the proposed approach in mitigating hallucinations and improving the recognition capability of LVLMs in diverse scenarios.

\section*{Acknowledgement}
This work is supported in part by the National Natural Science Foundation of China under grant 62171248, 62301189, 62576122, and Shenzhen Science and Technology Program under Grant KJZD20240903103702004, JCYJ20220818101012025, GXWD20220811172936001.

{
\small
\bibliographystyle{IEEEtran}
\bibliography{main}
}

\newpage
\section*{NeurIPS Paper Checklist}
\begin{enumerate}

\item {\bf Claims}
    \item[] Question: Do the main claims made in the abstract and introduction accurately reflect the paper's contributions and scope?
    \item[] Answer: \answerYes{}{} 
    \item[] Justification: We clearly state our contributions and research scope in the claims presented in the abstract and introduction, all of which are supported by theoretical foundation or extensive experiments.

\item {\bf Limitations}
    \item[] Question: Does the paper discuss the limitations of the work performed by the authors?
    \item[] Answer: \answerYes{} 
    \item[] Justification: We present the limitations of our paper in a section of the Appendix.

\item {\bf Theory assumptions and proofs}
    \item[] Question: For each theoretical result, does the paper provide the full set of assumptions and a complete (and correct) proof?
    \item[] Answer: \answerYes{} 
    \item[] Justification: We have provided full proof for the formulas in our paper.

    \item {\bf Experimental result reproducibility}
    \item[] Question: Does the paper fully disclose all the information needed to reproduce the main experimental results of the paper to the extent that it affects the main claims and/or conclusions of the paper (regardless of whether the code and data are provided or not)?
    \item[] Answer: \answerYes{} 
    \item[] Justification: Detailed information of experiments are provided in the Experimental section and the appendix.

\item {\bf Open access to data and code}
    \item[] Question: Does the paper provide open access to the data and code, with sufficient instructions to faithfully reproduce the main experimental results, as described in supplemental material?
    \item[] Answer: \answerYes{} 
    \item[] Justification: We submit our code in the supplementary material and will open-source the code, models, and data.

\item {\bf Experimental setting/details}
    \item[] Question: Does the paper specify all the training and test details (e.g., data splits, hyperparameters, how they were chosen, type of optimizer, etc.) necessary to understand the results?
    \item[] Answer: \answerYes{} 
    \item[] Justification: We provide exhaustive training and test details in our paper.

\item {\bf Experiment statistical significance}
    \item[] Question: Does the paper report error bars suitably and correctly defined or other appropriate information about the statistical significance of the experiments?
    \item[] Answer: \answerYes{} 
    \item[] Justification: We provide hyperparameter analysis in different settings.

\item {\bf Experiments compute resources}
    \item[] Question: For each experiment, does the paper provide sufficient information on the computer resources (type of compute workers, memory, time of execution) needed to reproduce the experiments?
    \item[] Answer: \answerYes{} 
    \item[] Justification: We provide them in the Appendix.
    
\item {\bf Code of ethics}
    \item[] Question: Does the research conducted in the paper conform, in every respect, with the NeurIPS Code of Ethics \url{https://neurips.cc/public/EthicsGuidelines}?
    \item[] Answer: \answerYes{} 
    \item[] Justification: The research conform every respect of the NeurIPS Code of Ethics.

\item {\bf Broader impacts}
    \item[] Question: Does the paper discuss both potential positive societal impacts and negative societal impacts of the work performed?
    \item[] Answer: \answerYes{} 
    \item[] Justification: We highlight that our work can reduce hallucinations in LVLMs and enhance their reliability in real-world applications.
    
\item {\bf Safeguards}
    \item[] Question: Does the paper describe safeguards that have been put in place for responsible release of data or models that have a high risk for misuse (e.g., pretrained language models, image generators, or scraped datasets)?
    \item[] Answer: \answerNA{} 
    \item[] Justification: The paper poses no such risks.

\item {\bf Licenses for existing assets}
    \item[] Question: Are the creators or original owners of assets (e.g., code, data, models), used in the paper, properly credited and are the license and terms of use explicitly mentioned and properly respected?
    \item[] Answer: \answerYes{}{} 
    \item[] Justification: The license and terms of use are properly respected.

\item {\bf New assets}
    \item[] Question: Are new assets introduced in the paper well documented and is the documentation provided alongside the assets?
    \item[] Answer: \answerNA{} 
    \item[] Justification: No new assets introduced except code.    

\item {\bf Crowdsourcing and research with human subjects}
    \item[] Question: For crowdsourcing experiments and research with human subjects, does the paper include the full text of instructions given to participants and screenshots, if applicable, as well as details about compensation (if any)? 
    \item[] Answer: \answerNA{} 
    \item[] Justification: No crowdsourcing nor research with human subjects.

\item {\bf Institutional review board (IRB) approvals or equivalent for research with human subjects}
    \item[] Question: Does the paper describe potential risks incurred by study participants, whether such risks were disclosed to the subjects, and whether Institutional Review Board (IRB) approvals (or an equivalent approval/review based on the requirements of your country or institution) were obtained?
    \item[] Answer: \answerNA{} 
    \item[] Justification: No crowdsourcing nor research with human subjects.

\item {\bf Declaration of LLM usage}
    \item[] Question: Does the paper describe the usage of LLMs if it is an important, original, or non-standard component of the core methods in this research? Note that if the LLM is used only for writing, editing, or formatting purposes and does not impact the core methodology, scientific rigorousness, or originality of the research, declaration is not required.
    \item[] Answer: \answerNA{} 
    \item[] Justification: We do not utilize LLMs for our core method development.

\end{enumerate}

\newpage
\appendix

\section{Complete Derivation of Equation (4)}
\label{app:proof}
Based on Eq. (\ref{equ:sentence_dis}) and Bayes' theorem, we provide the detailed derivation of Eq.~(\ref{equ:optim_expansion}) as follows:
\begin{align}
\max_{v, y} \mathrm{C\text{-}PMI}_{\theta}(v, y \mid x) &=\max_{v, y} \log \frac{p_{\theta}(v, y \mid x)}{p_{\theta}(v \mid x) \, p_{\theta}(y \mid x)} \\
&=\max_{v, y} \log \frac{p_{\theta}(v,x,y)/p_{\theta}(x)}{p_{\theta}(v \mid x) \, p_{\theta}(y \mid x)} \\
&=\max_{v, y} \log \frac{p_{\theta}(v,x,y)}{p_{\theta}(x)\, p_{\theta}(v \mid x)\, p_{\theta}(y \mid x)} \\
&=\max_{v, y} \log \frac{p_{\theta}(v,x,y)}{p_{\theta}(v, x) \, p_{\theta}(y \mid x)} \\
&=\max_{v, y} \log \frac{p_{\theta}(y\mid v,x)}{p_{\theta}(y \mid x)} \\
&=\max_{v, y} \log \frac{\prod_{t=0}^{l-1} {p}_{\theta}(y_t \mid v, x, y_{<t})}{\prod_{t=0}^{l-1} {p}_{\theta}(y_t \mid  x, y_{<t}))} \\
&=\max_{v, y} \log \prod_{t=0}^{l-1} {p}_{\theta}(y_t \mid v, x, y_{<t})-\log{\prod_{t=0}^{l-1} {p}_{\theta}(y_t \mid  x, y_{<t}))} \\
&=\max_{v, y}\sum_{t=0}^{l-1} \left[ \log p_{\theta}(y_t \mid v, x, y_{<t})- \log p_{\theta}(y_t \mid x, y_{<t}) \right].
\label{equ:proof_eq4}
\end{align}

\section{Experimental Details}
\label{sec:exp_details}

\subsection{Implementation Details}
Throughout our experiments, we retain 80\% of the visual input for LLaVA and LLaVA-NeXT, and 90\% for Shikra and InstructBLIP. To guide the training of the purifier, we utilize image-text pairs from ShareGPT4V \cite{chen2024sharegpt4v}—a high-quality image question answering dataset constructed using images from the MSCOCO dataset. Specifically, we use 2,000 samples for training the purifiers of LLaVA and LLaVA-NeXT, and 4,000 samples for InstructBLIP and Shikra. The learning rate is set to $1\times10^{-6}$ across all models for the decoding hyperparameters of LLMs, and the purifier is trained for 5 epochs. 
\subsection{Evaluation Model}
As mentioned above, we adopt InstructBLIP \cite{dai2023instructblip}, Shikra \cite{chen2023shikra}, LLaVA-1.5 on the 7B scale \cite{liu2024improved}, and LLaVA-NeXT \cite{li2024reqa} on the 8B scale. InstructBLIP employs Q-former as a cross-modal connector, leveraging 32 learned query tokens to extract and align visual features with text representations in an efficient manner. Other models adopt a simpler architecture of linear projection layers, which directly map visual features into the language model’s embedding space, typically using a larger number of image tokens (256 or even 576) as input.
\subsection{Evaluation Benchmarks}
\textbf{CHAIR Evaluations.} The Caption Hallucination Assessment with Image Relevance (CHAIR) metric is specifically designed to evaluate object hallucination in image captioning tasks. It quantifies the extent to which a generated caption includes references to objects that are not present in the corresponding ground-truth annotations. Specifically, CHAIR computes the proportion of hallucinated objects, those mentioned in the generated caption but absent from the reference object set, providing a direct measure of hallucination severity. CHAIR comprises two commonly used variants: CHAIR$_i$ ($C_I$) and CHAIR$_s$ ($C_S$), which evaluate the degree of object hallucination at the instance and sentence level, respectively. The lower values of $C_I$ and $C_S$ correspond to a lower degree of object hallucination, indicating greater factual consistency. The two variants can be formulated as follows:
$$
C_I=\frac{|\text{hallucinated\ objects}|}{|\text{all\ mentioned\ objects}|},\quad C_S=\frac{|\text{captions\ with\ hallucinated\ objects}|}{|\text{all\ captions}|}
$$
 
\textbf{POPE Evaluations.} The Polling-based Object Probing Evaluation (POPE) benchmark is also proposed to evaluate object hallucination in LVLMs. It adopts a discriminative approach by prompting models with binary questions such as “Is there a <object> in the image?” to assess whether the model can correctly identify the presence or absence of specific objects. To ensure balanced evaluation, POPE includes a 50\%/50\% ratio of queries about present and absent objects. POPE further categorizes the queries into three negative sampling settings: (1) \textit{random}, where absent objects are sampled randomly; (2) \textit{popular}, where negative objects are selected from the most frequent categories; (3) \textit{adversarial}, where negative objects are chosen based on their high co-occurrence likelihood with present ones to increase difficulty. Evaluation is conducted using Accuracy and F1 score, with higher scores indicating stronger performance in mitigating object hallucinations. Due to the concise format of POPE responses, which are typically short declarative sentences, the benchmark primarily reflects the visual grounding ability of a model rather than its long-form generation capacity.

\textbf{GPT-4 Assisted Evaluations.} In addition to object-level hallucinations via CHAIR and POPE, we adopt the GPT-4 assisted benchmark \cite{zhao2023beyond}, which leverages fine-grained object-level annotations from the Visual Genome (VG) dataset \cite{krishna2017visual} as ground truth. In our implementation, we employ the advanced GPT-4o to identify detailed hallucinations, such as positional, relational, and attribute-based errors, and compute the Sentence-level Hallucination Ratio (SHR) as evaluation results. Given the generated captions and manually annotated facts, GPT-4o is prompted by a template to assess hallucinations for every sentence. Following previous studies \cite{huang2024opera, huo2024self}, we evaluate on 200 VG images with a maximum output length of 512 tokens based on the prompt: "\texttt{Please describe this image in detail.}"

\textbf{MME and MMBench Evaluations.} MLLM Evaluation (MME) benchmark is designed to rigorously assess hallucination in MLLMs. It provides a suite of fine-grained, image-grounded multiple-choice questions across various categories, such as object recognition, OCR, counting, and commonsense reasoning, each requiring accurate visual understanding. By offering carefully controlled distractors and a consistent answer format, MME allows for precise evaluation of a model’s ability to generate faithful, image-grounded responses. MMBench is a large-scale, bilingual, multimodal benchmark designed to comprehensively evaluate the capabilities of vision-language models (VLMs). It consists of over 3,000 carefully curated multiple-choice questions covering 20 fine-grained ability dimensions, ranging from perception to reasoning. To ensure robustness and fairness, MMBench introduces the CircularEval strategy, where models must consistently answer a question across multiple permutations of choices. MME and MMBench provide rigorous and scalable frameworks for evaluating multimodal understanding and instruction-following capabilities across a wide spectrum of models.

\subsection{Principles of Hyperparameter Choices and Adaptation to New LVLMs.} 
\begin{figure*}[t]
\begin{minipage}[t]{0.47\linewidth}
  \centering
  \includegraphics[width=0.95\textwidth]{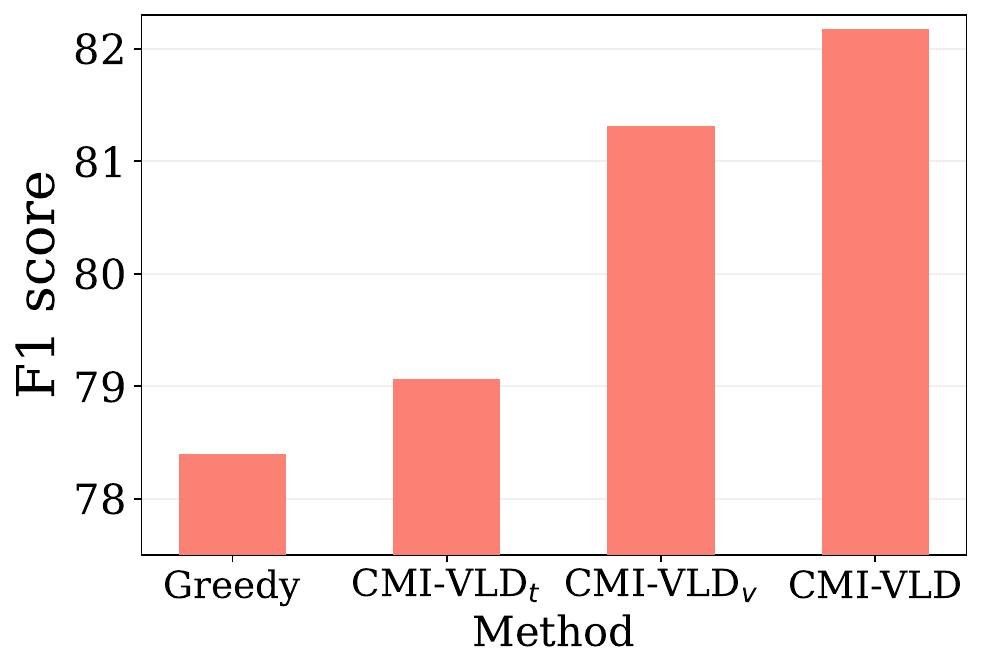}
  \centering
  \captionof{figure}{Ablation analysis of the proposed two techniques on the POPE metric.}
  \label{fig:ablation_two_techniques}
\end{minipage}
\begin{minipage}[t]{.06\linewidth}
\quad
\end{minipage}
\begin{minipage}[t]{.47\linewidth}
  \centering
  \includegraphics[width=0.95\textwidth]{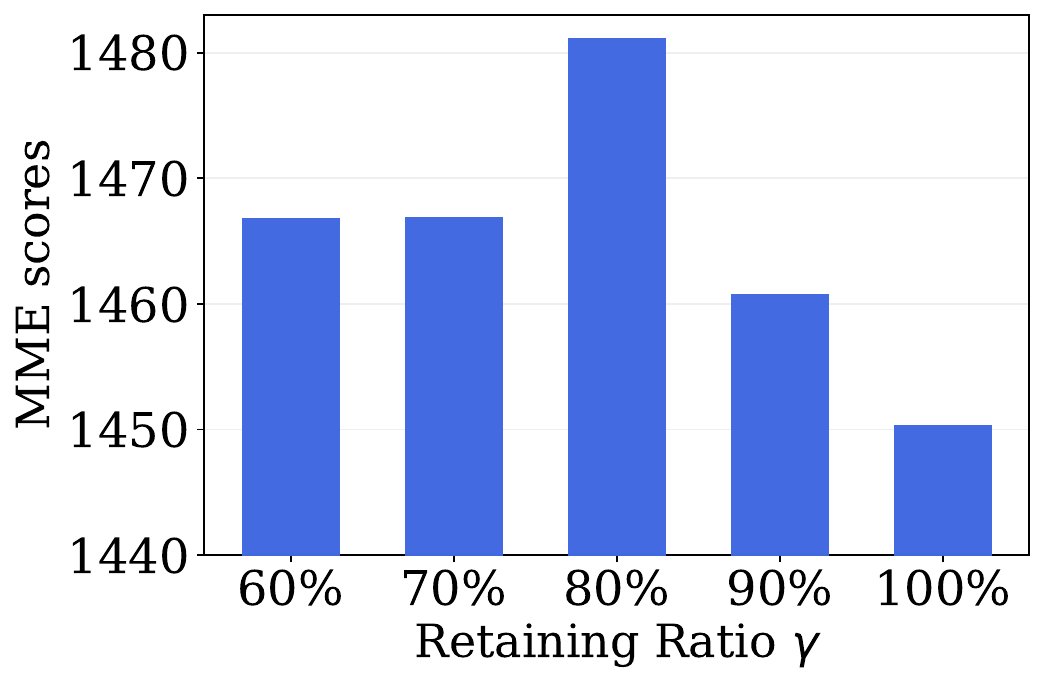}
  \captionof{figure}{MME results of the proposed CMI-VLD under varying values of retaining ratio $\gamma$.}
\label{fig:gamma}
\end{minipage}
\end{figure*}

The hyperparameters are chosen based on our empirical analysis and relevant literature, which are further supported through ablation studies.
Specifically, we summarize our choice strategy behind several critical hyperparameters  for LLaVA-1.5 in our algorithm as follows:
 \begin{itemize}
 
\item Contrast strength $\lambda$ balances the difference between the vision-conditioned and vision-free distributions. Large values may favor casual and incorrect tokens, while a small $\lambda$ can fail to sufficiently penalize hallucination-prone tokens.
We aim to preserve correct distributions while penalizing hallucinated predictions, and thus select a moderate value of $\lambda=0.5$, which is further validated by ablations (see Fig. \ref{fig:ablation_study} (b)) and prior studies \cite{huo2024self, leng2024mitigating}.

\item  Visual token retention ratio $\gamma$ is a sensitive and critical parameter. A high $\gamma$ may retain noisy tokens and weaken the C-PMI maximization, while a low can discard important visual information. Hence, we adopt an adaptive strategy: for models with many visual tokens (e.g., LLaVA-1.5), we set a relatively smaller $\gamma=80\%$; for models with fewer, already refined tokens (e.g., InstructBLIP), we use a higher $\gamma=90\%$. Ablation studies on each model validate the rationality, with results on LLaVA-1.5 shown in Fig. \ref{fig:gamma} as an illustration.

\item Coefficient of attention loss $\alpha$ balances the importance between attention loss and C-PMI loss during purifier training. Empirically, we find the attention loss to be $\sim$1000x smaller in magnitude compared to C-PMI loss, so we set $\alpha=100$
 to adequately amplify its impact while preserving the dominance of the C-PMI objective.

\item Purification starting layer $i$ is chosen based on existing well-established studies on token selection \cite{huo2024self, chen2024image}, which have been empirically validated to yield strong task-specific performance while preserving its general capability.
\end{itemize}
 
For the adaptation of our method, we empirically observe that the above hyperparameters can transfer to new LVLMs and achieve effectiveness. A necessary adaptation involves slightly tuning $\lambda$ according to the characteristics of the new models. We also recommend adjusting the retention ratio $\gamma$ based on the number of input visual tokens, as previously suggested.
In addition, it is also necessary to correspondingly adjust the hyperparameters in latent feature steering \cite{liu2025reducing} for different LVLMs.

\subsection{Explanation about the Inference Time Comparison}
Notably, following SID \cite{huo2024self}, we evict the selected visual tokens by applying masks to the attention matrices for implementation convenience, rather than physically removing them. The inference time of CMI-VLD reported in Fig.~\ref{fig:generation_time} is measured under this implementation and demonstrates that our method maintains inference efficiency. In practice, performing actual eviction of visual tokens would further accelerate inference, implying that the real efficiency advantage of CMI-VLD is even greater.

\begin{table}[t]
  \centering
  \caption{Comparison of the proposed CMI-VLD with SOTA baselines on the POPE metric. To make a fair comparison, all the methods are based on the \textit{sampling} decoding.}
    \resizebox{\linewidth}{!}{\begin{tabular}{cccccccc} 
    \toprule
    \multicolumn{1}{c}{\multirow{2}[0]{*}{Model}} & \multirow{2}[0]{*}{Method} & \multicolumn{2}{c}{Random} & \multicolumn{2}{c}{Popular} & \multicolumn{2}{c}{Adversarial} \\ \cmidrule(lr){3-4} \cmidrule(lr){5-6} \cmidrule(lr){7-8}
   \multicolumn{1}{c}{} &      & \textbf{Accuracy} & \textbf{F1 score} & \textbf{Accuracy} & \textbf{F1 score} & \textbf{Accuracy} & \textbf{F1 score} \\ \midrule
    \multirow{6}[0]{*}{LLaVA-1.5} & Default & 85.20\% & 85.42\% & 81.67\% & 82.50\% & 76.20\% & 78.40\% \\
          & ICD   & 85.73\% & 85.84\% & 81.90\% & 82.61\% & 76.70\% & 78.68\% \\
          & VCD   & 83.77\% & 84.24\% & 80.77\% & 81.84\% & 76.10\% & 78.38\% \\
          & VTI   & 85.23\% & 85.33\% & 82.77\% & 83.24\% & 76.63\% & 78.56\% \\
          & SID   & 87.93\% & 87.65\% & 84.57\% & 84.69\% & 79.43\% & 80.59\% \\
          & Ours  & \textbf{88.63\%} & \textbf{87.83\%} & \textbf{86.37\%} & \textbf{85.71\%} & \textbf{82.27\%} & \textbf{82.18\%} \\ \midrule
    \multirow{6}[0]{*}{Shikra} & Default & 85.07\% & 83.44\% & 83.13\% & 81.68\% & 81.63\% & 80.37\% \\
          & ICD   & 85.27\% & 83.87\% & 83.13\% & 81.94\% & 81.73\% & 80.73\% \\
          & VCD   & 85.17\% & 83.77\% & 83.27\% & 82.03\% & \textbf{82.03\%} & 80.96\% \\
          & VTI   & 84.03\% & 81.94\% & 82.43\% & 80.49\% & 81.07\% & 79.29\% \\
          & SID   & 85.53\% & 84.26\% & 83.47\% & 82.39\% & 81.43\% & 80.64\% \\
          & Ours  & \textbf{86.23\%} & \textbf{85.15\%} & \textbf{83.83\%} & \textbf{82.96\%} & 81.63\% & \textbf{81.08\%} \\ \bottomrule
    \end{tabular}}
  \label{tab:pope_more_models}%
\end{table}

\section{More Experimental Results}
\label{sec:more_exp}

\textbf{POPE evaluation on more LVLMs.} We supplement the results of POPE metrics on more LVLMs, including LLaVA-1.5 and Shikra. The quantitative results in Table \ref{tab:pope_more_models} again confirm the effectiveness of our method in mitigating object hallucinations.

\textbf{Ablation study of vision-language decoding.} We then conduct an ablation analysis to validate the contributions of the proposed two techniques, \textit{i.e.}, \textit{Calibrated Distribution Sampling} and \textit{Visual Token Refinement}, which interact with each other to fully maximize the C-PMI. Specifically, we design two variants CMI-VLD$_{t}$ and CMI-VLD$_{v}$, which retain only the \textit{Calibrated Distribution Sampling} and \textit{Visual Token Refinement}, respectively. Results in Fig. \ref{fig:ablation_two_techniques} reveal that both the removal of the two components degrade the performance of our algorithm, validating their considerable contributions to guarantee a successful approach for hallucination mitigation.

\textbf{Ablation study of varying retaining ratio.} The retaining ratio $\gamma$ is a sensitive hyperparameter that should be carefully tuned. A high retaining ratio may fail to sufficiently enhance C-PMI, whereas an excessively low value can degrade performance due to information loss. We evaluate the influence under varying $\gamma$ to confirm the optimal value. Fig. \ref{fig:gamma} indicates that $\gamma=80\%$ is an optimal choice.

\begin{table}[t]
  \centering
  \caption{Inference costs under varying sequence lengths. The number of visual tokens is fixed to 576.}
    \resizebox{0.92\linewidth}{!}{\begin{tabular}{ccccccc} \toprule
    \multirow{2}[0]{*}{Metric} & \multicolumn{1}{c}{\multirow{2}[0]{*}{Method}} & \multicolumn{5}{c}{Sequence Length} \\ \cmidrule(lr){3-7}
          &       & \multicolumn{1}{l}{633 (prefilling)} & 850   & 1000  & 1250  & 1500 \\  \midrule
    \multirow{2}[0]{*}{FLOPs (1e14)} & w/o purifier & 1.9214 & 2.5814 & 3.0391 & 3.8044 & 4.5731 \\
          &  CMI-VLD & 1.5671 & 2.4582 & 3.1226 & 4.3182 & 5.6241 \\ \midrule
    \multirow{2}[0]{*}{Inference Latency (s)} & w/o purifier & 0.37  & 17.99 & 30.26 & 50.43 & 70.72 \\
          &  CMI-VLD & 0.35  & 18.53 & 31.17 & 52.2  & 73.22 \\ \bottomrule
    \end{tabular}}
  \label{tab:cost_seq}%
\end{table}%
\begin{table}[t]
  \centering
  \caption{Inference costs under varying numbers of input visual tokens. We use a fixed text query from the CHAIR evaluation, where the number of text tokens is 56.}
    \resizebox{0.76\linewidth}{!}{\begin{tabular}{cccccc} \toprule
  \multirow{2}[1]{*}{Metric} & \multicolumn{1}{c}{\multirow{2}[1]{*}{Method}} & \multicolumn{4}{c}{Visual Token Count} \\ \cmidrule(lr){3-6}
          &       & 49    & 256   & 576   & 1024 \\ \midrule
    \multirow{2}[0]{*}{FLOPs (1e14)} & w/o purifier & 0.3188 & 0.9448 & 1.9214 & 3.3067 \\
          &  CMI-VLD & 0.2888 & 0.7876 & 1.5671 & 2.6718 \\ \midrule
    \multirow{2}[0]{*}{Inference Latency (s)} & w/o purifier & 0.24  & 0.28  & 0.37  & 0.60 \\
          & CMI-VLD & 0.23  & 0.28  & 0.35  & 0.53 \\ \bottomrule
    \end{tabular}}
  \label{tab:cost_visual}%
\end{table}%

\textbf{Detailed Analysis of the Computational Costs.}
To reduce computational overhead, we design the purifier as a lightweight network with only ~0.1\% of the parameters of the LVLM. Its effectiveness in mitigating hallucination has been thoroughly validated by extensive experiments in the main text. Next, we present a detailed computational cost analysis of the visual purifier using LLaVA-Next 8B.

As shown in Table \ref{tab:cost_seq} and \ref{tab:cost_visual}, thanks to its lightweight design and visual token reduction, our purifier introduces negligible overhead and generally maintains computational efficiency comparable to the purifier-free variant. Furthermore, as the number of visual tokens increases, the benefits of visual token reduction become more pronounced, further reducing the computational complexity.

\begin{table}[htbp]
  \centering
  \caption{CHAIR metrics and Token-per-second (TPS) of CMI-VLD and its learning-free variant CMI-VLD$_{lf}$ on four LVLMs using greedy decoding. We present the results of the existing SOTA method, SID \cite{huo2024self}, as a reference.}
    \resizebox{0.7\linewidth}{!}{\begin{tabular}{cccccc} \toprule
    \multicolumn{1}{l}{Metric} & Method & \multicolumn{1}{l}{LLaVA-1.5} & \multicolumn{1}{l}{InstructBLIP} & \multicolumn{1}{l}{Shikra} & \multicolumn{1}{l}{LLaVA-NEXT} \\ \midrule
    \multirow{3}[0]{*}{C$_S\downarrow$} & SID   & 42.8  & 56.2  & 51.2  & 38.0 \\
          & CMI-VLD$_{lf}$ & 30.0  & \textbf{40.4} & 36.2  & \textbf{26.6} \\
          & CMI-VLD & \textbf{29.9} & 43.2  & \textbf{30.6} & 27.2 \\ \midrule
    \multirow{3}[0]{*}{C$_I\downarrow$} & SID   & 12.1  & 15.8  & 13.6  & 8.9 \\
          & CMI-VLD$_{lf}$ & 9.0   & \textbf{11.8} & 10.2  & \textbf{6.5} \\
          & CMI-VLD & \textbf{8.9} & 12.9  & \textbf{8.9} & 6.8 \\ \midrule
    \multirow{3}[0]{*}{TPS $\uparrow$} & SID   & 8.76  & 11.70 & 3.85  & 15.71 \\
          & CMI-VLD$_{lf}$ & 2.45  & 2.41  & 1.05  & 2.35 \\
          & CMI-VLD & \textbf{8.96} & \textbf{11.86} & \textbf{4.29} & \textbf{16.52} \\ \bottomrule
    \end{tabular}}
  \label{tab:ablation_lf}%
\end{table}%

\textbf{Investigation of the learning-free variant.} 
Initially, we proposed learning a purifier to reduce the computational overhead incurred by manual token selection. To validate this design, we implement a learning-free variant CMI-VLD$_{lf}$
, which selects tokens by manually computing our derived score in Eq. (7) at each step, with all other settings unchanged.

Table \ref{tab:ablation_lf} shows that both variants of our method significantly mitigate hallucination compared to the SOTA baseline, validating the effectiveness of our objective function derived from C-PMI. However, manual token selection incurs substantial latency due to repeated score computations at each decoding step, limiting its practicality in real-world applications.
In contrast, our learned purifier efficiently selects informative tokens with nearly 4× faster inference than CMI-VLD$_{lf}$
 while preserving strong effectiveness, exhibiting an excellent trade-off between performance and efficiency.

\section{Model Architecture of Visual Token Purifier}
\label{sec:architecture_purifier}
We provide the detailed purifier architecture as follows. Notably, this learnable network contains fewer than 1\% of the LVLM's total parameters, hence introducing only marginal computation overheads.
\begin{figure}[htbp]
\centering
\includegraphics[width=0.95\linewidth]{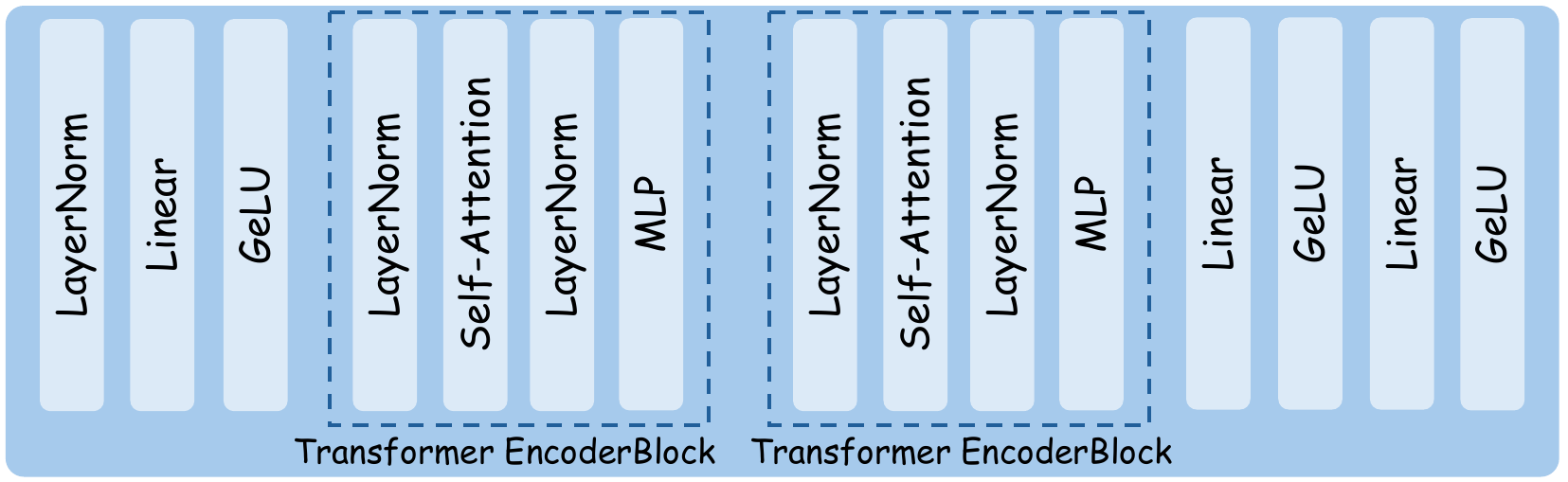}
\caption{Overview of the architecture of the visual token purifier.}
\label{fig:purifier}
\end{figure}

\section{Limitations}
Despite the promising performance of LVLMs, the proposed method still faces several limitations. First, the introduced visual purifier introduces additional computational overhead for purifier training. Second, when the LVLM generates very long responses, the efficiency gain from the removal of visual tokens may become less significant. Besides, the performance of the LVLM is highly sensitive to the retaining ratio, and the optimal ratio may vary at each decoding step. However, our method adopts a fixed mask rate throughout the generation process. Future work could better address this limitation by exploring more advanced purification strategies with adaptive retaining ratios.

\section{Visualization Results}
\begin{figure}[htbp]
\centering
\includegraphics[width=0.95\linewidth]{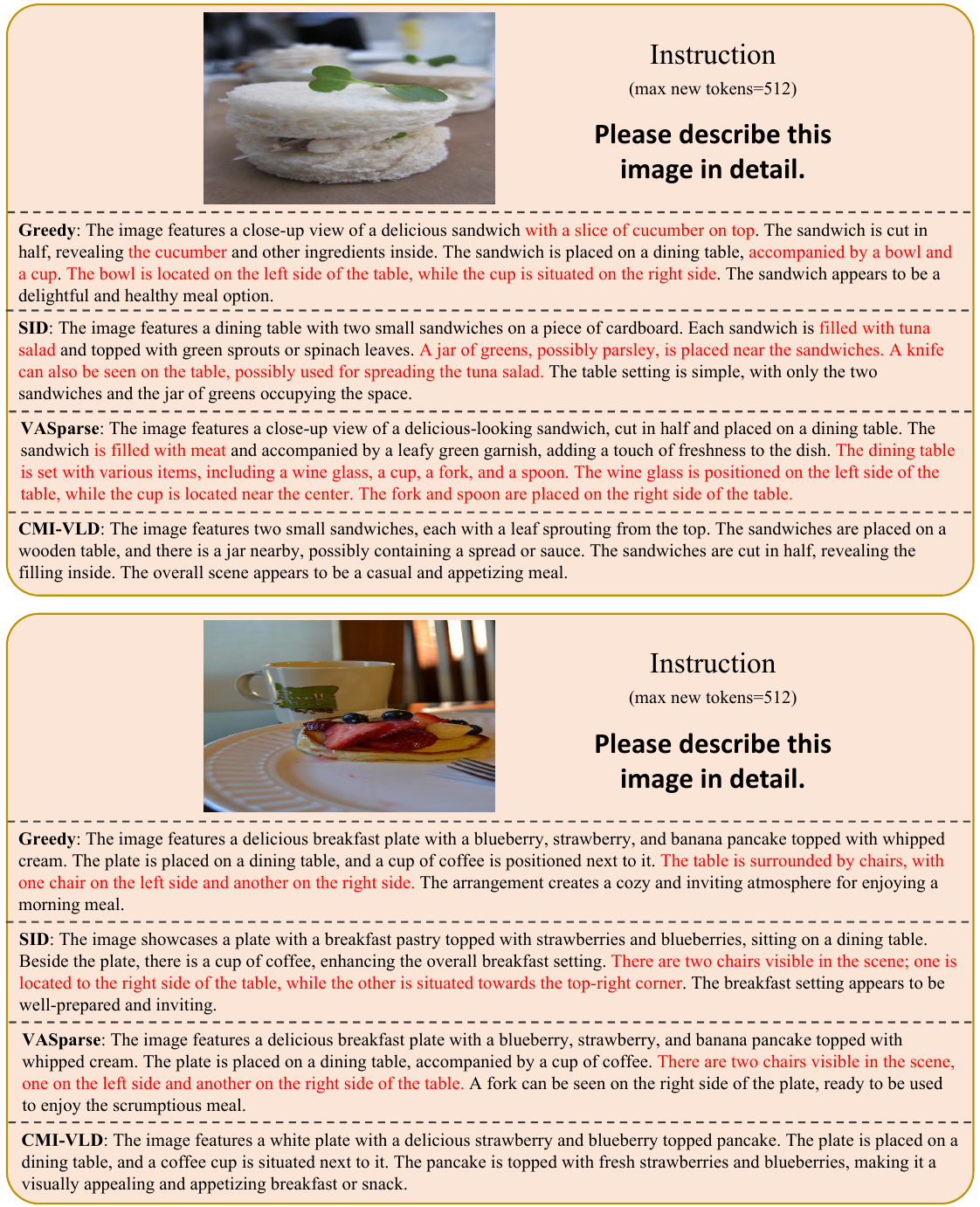}
\caption{Visualization results comparing our CMI-VLD and other methods with LLaVA-1.5 backbone. Hallucinations are marked in \textcolor{red}{red}.}
\label{fig:visualization}
\end{figure}

\begin{figure}[htbp]
\centering
\includegraphics[width=\linewidth]{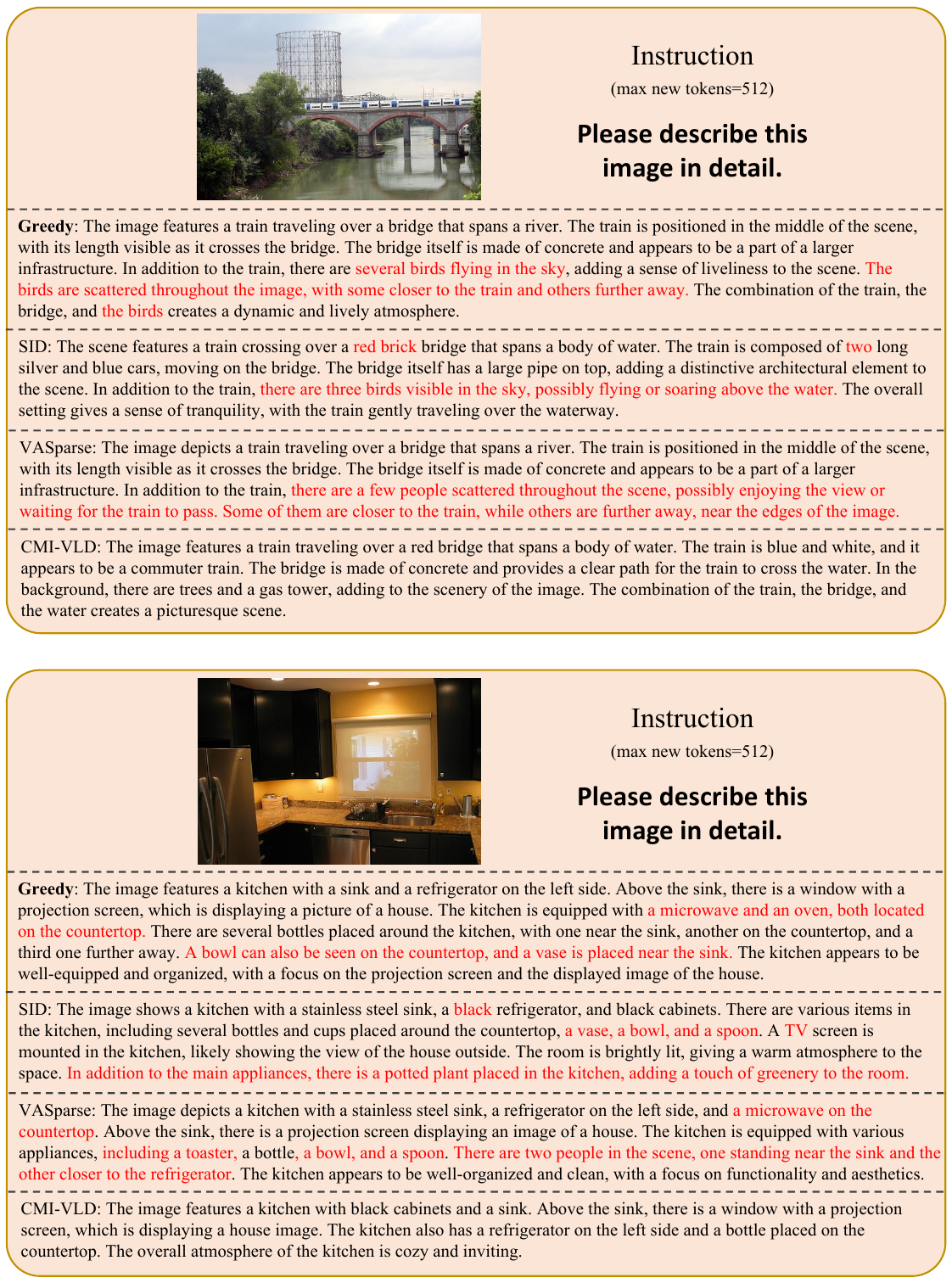}
\caption{Visualization results comparing our CMI-VLD and other methods with LLaVA-1.5 backbone. Hallucinations are marked in \textcolor{red}{red}.}
\label{fig:visualization_2}
\end{figure}

To visually demonstrate the effectiveness of our approach, we present generated captions using the prompt "\texttt{Please describe this image in detail.}" from our method and baseline approaches on the MSCOCO dataset in Fig. \ref{fig:visualization} and \ref{fig:visualization_2}. Qualitative results reveal the effectiveness of our method.

\label{sec:more_visualization}

\end{document}